# Wavelet based approach for tissue fractal parameter measurement: Pre cancer detection

Sabyasachi Mukhopadhyay[1], Nandan K. Das[1], Soham Mandal[2], Sawon Pratiher[3], Asish Mitra[4], Asima Pradhan[3], Nirmalya Ghosh[1], Prasanta K. Panigrahi[1]

[1] Indian Institute of Science Education and Research (IISER) Kolkata, Nadia-741246, India

[2] Institute of Engineering & Management, Kolkata

[3] Indian Institute of Technology (IIT) Kanpur, Kanpur-208016, India

[4] College of Engineering & Management, Kolaghat, India

## ABSTRACT

In this paper, we have carried out the detail studies of pre-cancer by wavelet coherency and multifractal based detrended fluctuation analysis (MFDFA) on differential interference contrast (DIC) images of stromal region among different grades of pre-cancer tissues. Discrete wavelet transform (DWT) through Daubechies basis has been performed for identifying fluctuations over polynomial trends for clear characterization and differentiation of tissues. Wavelet coherence plots are performed for identifying the level of correlation in time scale plane between normal and various grades of DIC samples. Applying MFDFA on refractive index variations of cervical tissues, we have observed that the values of Hurst exponent (correlation) decreases from healthy (normal) to pre-cancer tissues. The width of singularity spectrum has a sudden degradation at grade-I in comparison of healthy (normal) tissue but later on it increases as cancer progresses from grade-II to grade-III.

## 1. INTRODUCTION

Biological samples are the most complex structures in nature. The biochemical information on this sample can be exploited for probing subtle biochemical changes as a signature of disease progression. The morphological approach provides us functional information of potential biomedical importance. The refractive index fluctuations in biological samples are fractal in nature which can be exploited to understand the changes in morphology through the progress of various diseases. There are complex interactions between neoplastic cells and the stromal during progression of cancer. Also, carcinogenesis results, in part, from defective epithelial-stromal communication [1]. Interestingly it can be observed that the collagen fiber network present in stromal also exhibits fractal architecture in the organization of the fibers and micro-fibrils [2]. The MFDFA is a very powerful statistical tool capable of detecting hidden long range correlations in noisy, non-stationary, multifractal fluctuation series, and has been successfully established tool in diverse fields [3]. The light scattering models for quantification of the fractal micro-optical properties, namely, Hurst exponent (H) and fractal dimension ($D_f$), have been explored for their potential applications in precancer detection of recent studies [4-7]. We know that wavelet transform is a very powerful tool for data analysis. The ability of the wavelets to provide multiresolution, in addition to their localization properties which makes them an ideal tool for studyingdata sets, having different structures. The wavelet basis can be used for effectively capture of collectivebehavior and sharp changes and for localizing them [11-12]. Furthermore the mathematical microscopic nature of the wavelets can also be used for analyzing the localized performance at various scales [19-21]. Using wavelets, analysis of a number of data sets of both parallel and perpendicular polarized spectra has been done which have led to several key distinctions between different tumors and corresponding normal breast tissues. It has revealed the usefulness of polarized fluorescence in the diagnosis of tumors [18]. Quantification of these differences and fluctuations in the perpendicular channel of the cancerous tissues have been found more randomized as compared to their normal counterparts using wavelets [17]. In this paper, wavelet transform has been applied to extract some features of normal and different grades of DIC images. MFDFA has been applied to show how Hurst exponent (H) and fractal dimension ($D_f$) differs from normal to different grades of cancer of stromal regions.

## 2. THEORY

### 2.1 WAVELET TRANSFORM (WT)

For a given continuous time signal $x(t)$, the wavelet transform is defined as:

$$T(a,b) = \frac{1}{\sqrt{a}} \int_{-\infty}^{\infty} x(t)\psi^*(\frac{t-b}{a})dt \qquad (1)$$

Here $\psi^*(t)$ is the complex conjugate of the wavelet function of $\psi(t)$ and $a$, $b$ are the dilation parameter and the location parameter of the wavelet respectively. A wavelet needs to satisfy (i) admissibility condition and (ii) should have finite energy: $E = \int_{-\infty}^{\infty} |\psi(t)|^2 dt < \infty$ [9].

If the low pass and high pass coefficients are $c_k$'s and $d_{j,k}$'s respectively, the data set can be expressed for Discrete wavelet transform as $f(t) = \sum_k c_k \phi_k + \sum_k \sum_{j=0}^{\infty} d_{j,k} \psi_{j,k}$. The trend components are extracted by father wavelet $\phi_k$, located at $k$ with level $j$ and the deviations from the trend are picked up by mother wavelets $\psi_{j,k}$ [15].

### 2.2 WAVELET COHERENCE

Coherence is considered to be equivalent to correlation. Though, there are important differences between them. In coherence's calculation the signal is squared, thus producing values from 0 to 1. The polarity information is lost. By contrast, correlation is sensitive to polarity and its values range from -1 to 1. The wavelet-based estimation of coherence, or wavelet coherence, is a very recent tool. Wavelet coherence can be written as: $\rho(a,s) = \frac{SW_{xy}^2(a,s)}{SW_{xx}(a,s)SW_{yy}(a,s)}$. The computation of the wavelet coherence value requires the exact values of the wavelet auto- and wavelet cross-spectra of x and y as: $SW_{xy}(a,s) = W_x(a,s)W_y^*(a,s)$.

Where $W_x(a,s)$ is the wavelet coefficient at scale $s$ and time $t$ of the finite time series $x(t)$, and $W_y^*(a,s)$ the complex conjugate of $W_y(a,s)$. Wavelet coherence phase are defined as:

$$PC(f,t) = \frac{1}{\delta} \int_{t-\frac{\delta}{2}}^{t+\frac{\delta}{2}} \frac{W_x(\tau)W_y^*(\tau)}{|W_x(\tau)W_y^*(\tau)|} d\tau, \text{ where } \frac{W_x(\tau)}{W_y(\tau)} = \exp(j(\varphi_y(f,\tau))) \text{ and } \tau \text{ is a phase of signal } \phi(\tau).$$

### 2.3 MULTI FRACTAL DETRENDED FLUCTUATION ANALYSIS (MFDFA)

The mathematical details of MFDFA can be found reference [14]. Briefly, the profile $Y(t)$ (spatial series of length N, $t = 1,....,N$) is first generated from the one dimensional spatial index fluctuations. The profile is then divided into $N_s$ = int (N/s) non-overlapping segments $b$ of equal length s. The local trend of the series ($y_b(i)$) is determined for each segment $b$ by least square polynomial fitting, and then subtracted from the segmented profiles to yield the de-trended fluctuations. The resulting variance of the de-trended fluctuation is determined for each segment as

$$F^2(b,s) = \frac{1}{s}\sum_{i=1}^{s}\left[Y\{(b-1)s+i\} - y_b(i)\right]^2 \qquad (2)$$

The moment (q) dependent fluctuation function is then extracted by averaging over all the segments as

$$F_q(s) = \left\{\frac{1}{2N_s}\sum_{b=1}^{2N_s}[F^2(b,s)]^{\frac{q}{2}}\right\}^{1/q} \qquad (3)$$

The scaling behavior is subsequently determined by analyzing the variations of $F_q(s)$ vs s for each values of q, assuming the general scaling function as

$$F_q(s) \sim s^{h(q)} \qquad (4)$$

Here, the generalized Hurst exponent h(q) and the classical multifractal scaling exponent $\tau(q)$ are related by

$$\tau(q) = qh(q) - 1 \qquad (5)$$

Note that for a stationary, monofractal series $h(q=2)$ is identical to the Hurst exponent H [14]. Here, values of H = 0.5, > 0.5 and < 0.5 correspond to uncorrelated random fluctuations, long range correlated and anti-correlated behaviour respectively [14]. The two sets of the scaling exponents $h(q)$ and $\tau(q)$ along with the singularity spectrum $f(\alpha)$ completely characterize any non-stationary, multifractal fluctuation series. Here, $f(\alpha)$ is related to $\tau(q)$ via a Legendre transformation:

$$\alpha = \frac{d\tau}{dq}, \quad f(\alpha) = q\alpha - \tau(q) \qquad (6)$$

Where $\alpha$ is the singularity strength and the width of $f(\alpha)$ is a quantitative measure of multifractality[3].

## MATERIALS & METHODS

Here the spatial distribution of tissue refractive index is measured by the differential interference contrast (DIC) microscope (Olympus IX81, USA). The biopsy samples of human cervical tissues with total of 94 samples (26 normal, 26 grade I, 19 grade II, 23 grade III), which were histopathologically characterized, were obtained from G.S.V.M. Medical College and Hospital, Kanpur, India. The pixel-wise unfolding was done for recorded images in one linear direction for MFDFA analysis purpose for representing the spatial variation of tissue refractive index. Wavelet and MFDFA analysis were used for DIC image analysis purpose. In this current paper, the results of several sets of samples consisting of Normal to Grade-3 tissue images are shown. The DIC sample images are given below:

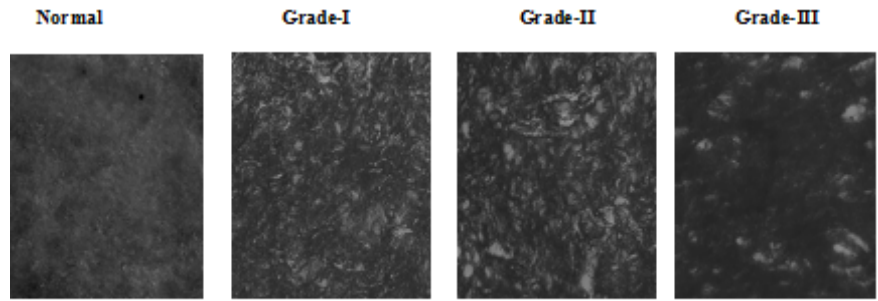

## RESULTS AND DISCUSSIONS

We know that the DIC images are useful for extracting tissue refractive index fluctuations. So, DIC image data can be considered as irregular signals. Wavelets are ideal analysis tool irregular data processing purpose[26]. For any observed signal x(t)=f(t)+e(t), where f(t) is the signal and e(t) is the noise, using wavelets f(t) can easily be extracted out. In this paper, Discrete wavelet transform (DWT) through Daubechies basis analysis has been done up to level-5 for identifying localized fluctuations (high pass coefficients) over polynomial trends for clear characterization and differentiation of tissues. It is clear that the high pass coefficients retain the refractive index fluctuations of the DIC images.

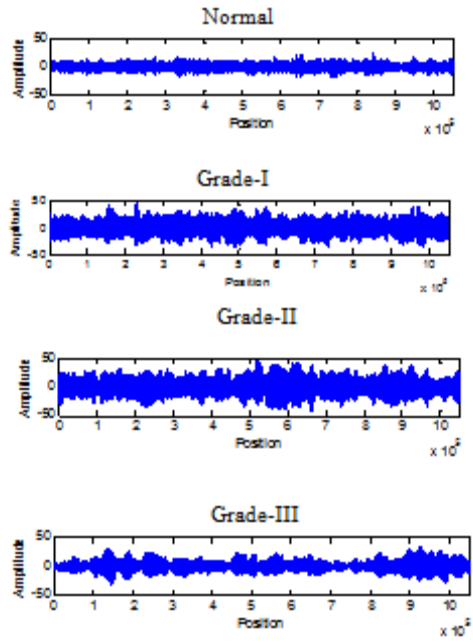

Fig1- Fluctuations DIC signals after 5-level decomposition

Here it is clearly visible that after level-5 decomposition amplitude peak values are more for Grade-I to Grade-II than the normal DIC images. At Grade-III again the amplitude values decreases. It occurs due to the refractive index fluctuations of the medium.

Thereafter, the wavelet coherence plots are done to check auto-correlation within localized region for analyzing the fluctuations of various samples by using complex Gaussian wavelet. In the wavelet coherence plots, the arrows are used in the figure to represent the relative phase between the two DIC images as a function of scale and position. As the relative phase information produces a local measure of the delay between the two DIC images, the wavelet coherence is superimposed with these relative phase plots. Here wavelet coherence plots are performed to show the inter-correlation between two samples of DIC normal and cancerous tissues.

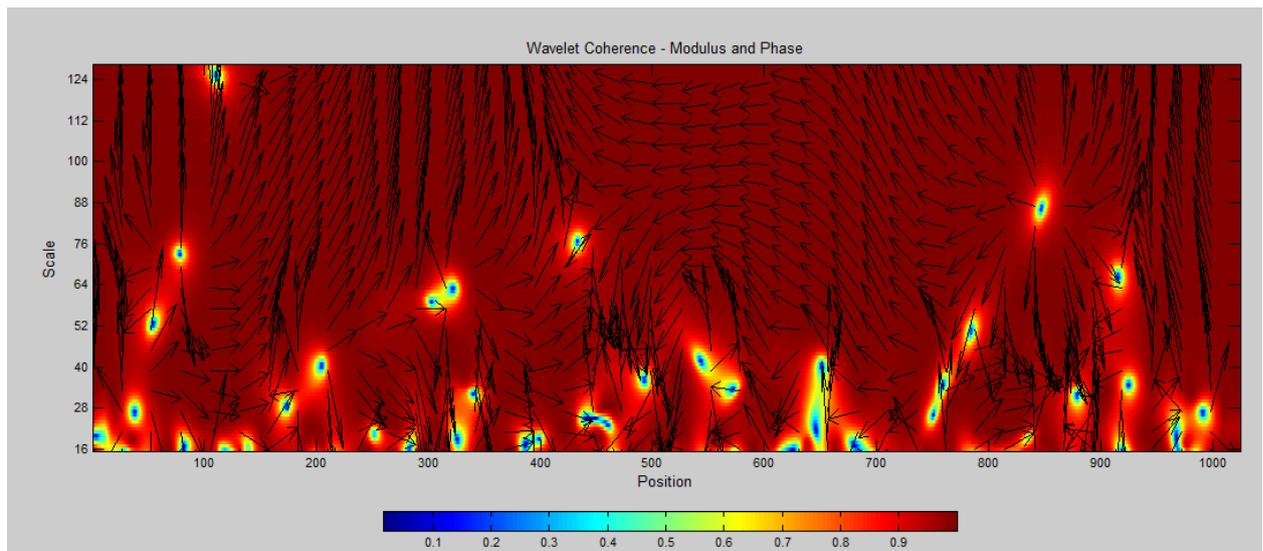

Fig.4a-Wavelet Coherence between Normal DIC and Grade-I Image

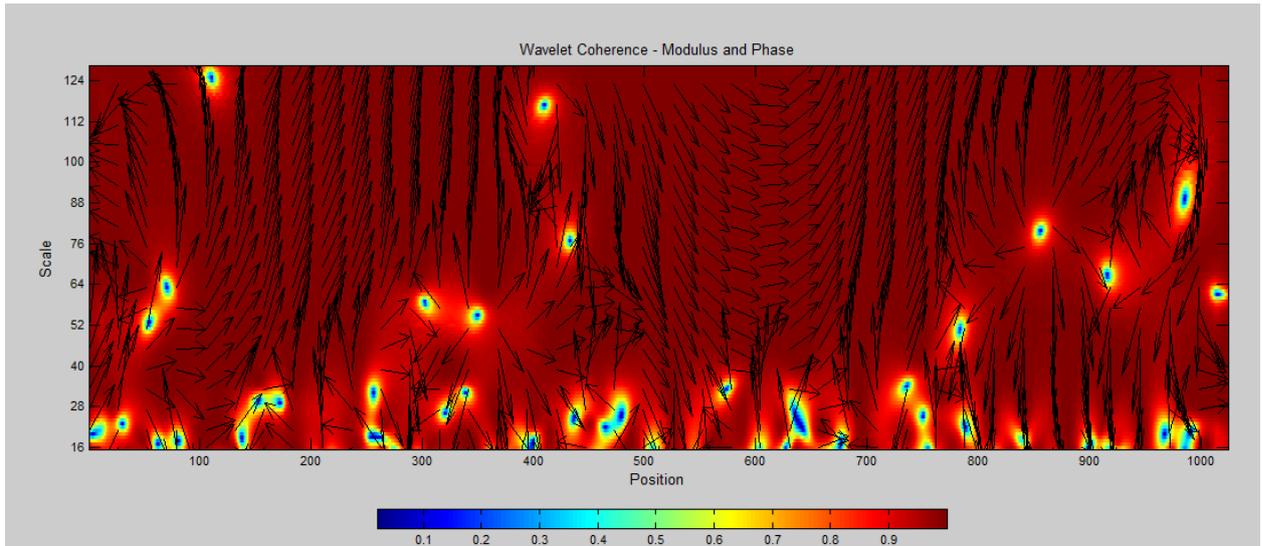

Fig.4b-Wavelet Coherence between Normal DIC and Grade-II Image

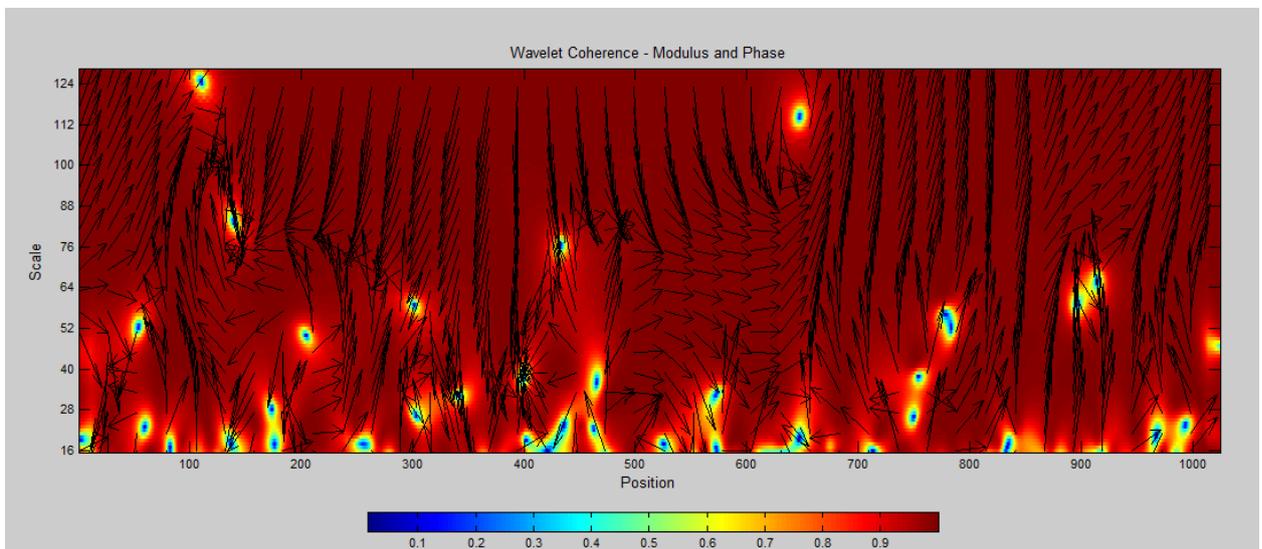

Fig.4c-Wavelet Coherence between Normal DIC and Grade-III Image

Here it appears that there are several distinct zones of low correlation between normal and cancerous tissues due to refractive index fluctuations. The regions with low coherency, relative phases (as shown by arrow) between normal and different grades are jumbled. It can be also observed that the jumbled phases are more prominent between normal and grade-III. From above plots we notified the highly coherent zones of various cancerous tissues with normal tissue.

In our previous work we showed during multifractal analysis how the value of Hurst exponent and Singularity spectrum varies among various cancerous tissues for epithelium region[8,25]. In this paper, the MFDFA analysis is done for stromal region.

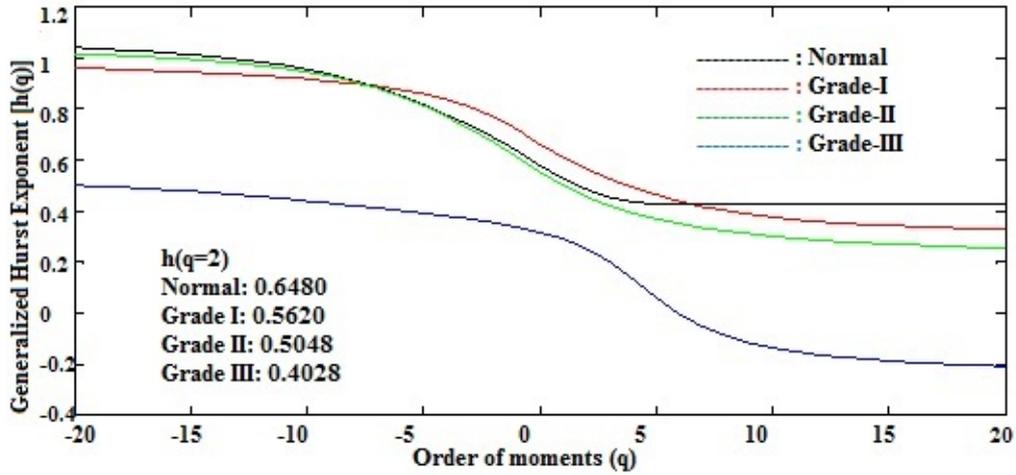

Fig.5a- Plot of Hurst Exponent from Normal to Grade-3 DIC Image

The Hurst exponent values of several samples are given in a tabular format in detail. It is clearly observed that the hurst exponent value decreases from normal to cancerous tisses due to refractive index fluctuations of the medium.

Table 1- Hurst Exponent Calculation using MFDFA

|  | Normal | Grade-I | Grade-II | Grade-III |
|---|---|---|---|---|
| Hurst Exponent(mean $h(q=2) \pm$ standard deviation) | $0.6480 \pm 0.02$ | $0.5620 \pm 0.06$ | $0.5048 \pm 0.01$ | $0.4028 \pm 0.07$ |

The singularity spectrum plot is shown below.

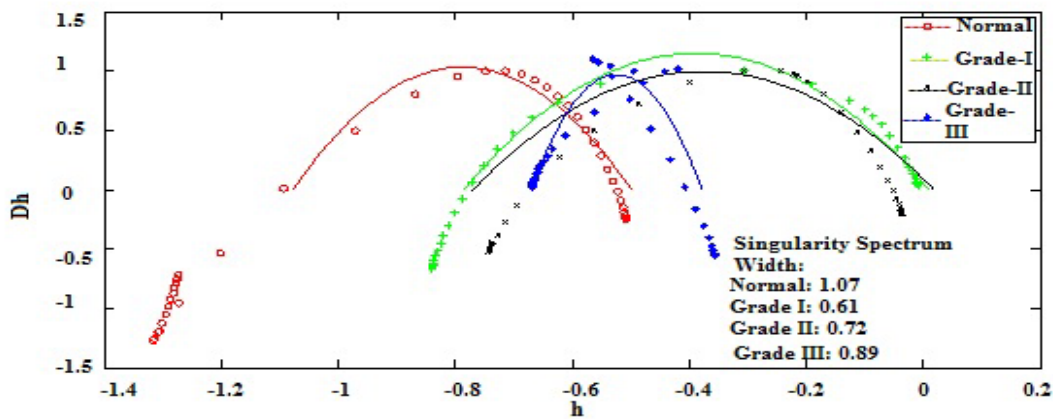

Fig.5b- Singularity Spectrum for DIC Image

The Singularity spectrum values of several samples are given in a tabular format in detail.

Table 2- Singularity Spectrum Width Calculation using MFDFA

|  | Normal | Grade-1 | Grade-2 | Grade-3 |
|---|---|---|---|---|

| Width of Singularity Spectrum(mean $h(q = 2)$ ± standard deviation) | 1.07 ± 0.03 | 0.61 ± 0.07 | 0.72 ± 0.09 | 0.8995 ± 0.05 |

The singularity width decreases from normal to grade-I and gradually increaes among different cancerous grades due to refractive index fluctuations.

## CONCLUSIONS

From the above results and discussions, it is clear that wavelet coherency and MFDFA are playing a vital role to show the correlation among healthy and different grades of pre-cancer tissues in time scale plane as well as characterization between them respectively. The authors hope that the above findings will definitely help researchers in future to make step forward in Bio-medical field.

**Acknowledgments**

This work was supported by Indian Institute of Science Education and Research (IISER)-Kolkata, an autonomous teaching and research institute funded by the Ministry of Human Resource Development (MHRD), Govt. of India. Author N. K. Das acknowledge to CSIR for providing fellowship to carryout research work.

## References

1. S. M. Pupa, S. M´enard, S. Forti, E. Tagliabue, "New insights into the role of extracellular matrix during tumor onset and progression," J. Cell.Physiol.,vol. 192, no. 3, pp. 259–267 (2002).

2. D. Arifler, I.Pavlova, A.Gillenwater, R. Richards-Kortum,"Light scattering from collagen fiber networks: Micro-optical properties of normal and neoplastic stroma", Biophysical Journal, vol. 92, no.9, pp. 3260 – 3274 (2007).

3. J. W. Kantelhardt, S. A. Zschiegner, E. Koscielny-Bunde, S. Havlin, A.Bunde, S. Havlin, H. E. Stanley, "Multifractal detrended fluctuation analysis of nonstationary time series", Physica A, **316**, pp. 87-114 (2002).

4. M. Hunter, V. Backman, G. Popescu, M. Kalashnikov, C. W. Boone, A. Wax, V. Gopal, K. Badizadegan, G. D. Stoner, M. S. Feld, "Tissue self-affinity and polarized light scattering in the Born approximation: A new model for precancer detection", Physical Review Letters, **97**, 138102 (2006).

5. M. Xu, R. R. Alfano, "Fractal mechanisms of light scattering in biological tissue and cells", Optics Letters, **30**, pp. 3051-3053 (2005).

6. C. J. R. Sheppard, "Fractal model of light scattering in biological tissue and cells", Optics Letters**, 32**, pp.142 -144 (2007).

7. I. R. Capoglu, J. D. Rogers, A. Taflove, V. Backman, "Accuracy of Born approximation in calculating the scattering coefficient of biological contenuous random medium", Optics Letters, **34**, pp.2679 – 2681 (2009).

8. N.Das, S.Chatterjee, A.Pradhan, P.K.Panigrahi, I. A.Vitkin, and N.Ghosh., "Tissue multifractality and Born approximation in analysis of light scattering: a novel approach for precancers detection", Nature Scientific Report, (2014).

9. I. Daubechies, Ten Lectures on Wavelets. Philadelphia, PA: Society for Industrial and Applied Mathematics, vol. 64, 1992, CBMS-NSF Regional Conference Series in Applied Mathematics.

10. N.Das, S Chatterjee, S Chakraborty, PK Panigrahi, A Pradhan, N Ghosh, "Fractal anisotropy in tissue refractive index fluctuations: potential role in pre-cancer detection", Proc. of SPIE, SPIE BIOS,Vol 9129, 91290V-1 (2014).

11. N.Agarwal, S.Gupta, Bhawna, A.Pradhan, K. Vishwanathan, P.K. Panigrahi," Wavelet Transform of Breast Tissue Fluorescence Spectra: A Technique for Diagnosis of Tumors ", IEEE J. Quantum Electron., Vol. 9,pp. 154-161, (2003).

12. S.G, MALLAT, " A WAVELET TOUR OF SIGNAL PROCESSING", 2ND ED.; ACADEMIC PRESS: ORLANDO, FL, USA, (1998).

13. N.Thekkek, R. Richards-Kortum, "Optical imaging for cervical cancer detection: solutions for a continuing global problem", Nature Review Cancer, **8**, pp. 725 – 731 (2008).

14. R. R Alfano, B. B. Das, J. Cleary, R. Prudente, E. Celmer, "Light sheds light on cancer distinguishing malignant tumors from benign tissues and tumors," Bull. NY Acad. Med., 2nd Series, vol. 67, pp. 143-150 (1991).

15. J.K. Modi, S.P. Nanavati, A.S. Phadke, P.K. Panigrahi, "Wavelet Transforms: Application to Data Analysis - II", Resonance, pp.8-13 (2004).

16. S. Mukhopadhyay, N. Das, A. Pradhan, N. Ghosh, P.K. Panigrahi, "Pre-cancer Detection by Wavelet Transform and Multi-fractality in various grades of DIC Stromal Images", Proc. of SPIE, SPIE BIOS, USA, pp.89420H.1-6(2014).

17. S.Gupta, M.S.Nair, A.Pradhan, N.C.Biswal, N.Agarwal, A.Agarwal, P.K., "Wavelet-based characterization of spectral fluctuations in normal, benign, and cancerous human breast tissues", Journal of Biomedical Optics, Vol-10(5), (2005).

18. N.Agarwal, S.Gupta, Bhawna, A.Pradhan, K.Vishwanathan, Prasanta K. Panigrahi, "Wavelet Transform of Breast Tissue Fluorescence Spectra: A Technique for Diagnosis of Tumors", IEEE JOURNAL OF SELECTED TOPICS IN QUANTUM ELECTRONICS, VOL. 9, NO. 2, MARCH/APRIL 2003

19. C. Chui, An Introduction to Wavelets, Academic Press, New York (1992).
20. G. Kaiser, A Friendly Guide to Wavelets, Birkhäuser, Boston, MA (1994).
21. G. W. Wornell, Signal Processing with Fractals: A Wavelet Based Approach, Prentice Hall, Englewood Cliffs, NJ (1996).
22. J. Lackowicz, Principles of Fluorescence Spectroscopy, Plenum Press, New York (1983).
23. R. R. Alfano, G. C. Tang, A. Pradhan, W. Lam, D. S. J. Choy, E.Opher, "Fluorescence spectra from cancerous and normal human breast and lung tissues," IEEE J. Quantum Electron. **23**, 1806–1811 (1987).
24. N. Ramanujam, M. F. Mitchell, A. Mahadevan-Jansen, S. Thomsen, G. Staerkel, A. Malpica, T. Wright, A. Atkinson, R. Richards-Kortum, "Cervical pre-cancer detection using a multivariate statistical algorithm based on laser induced fluorescence spectra at multiple excitation wavelengths," Photochem. Photobiol.**64**, 720–735 (1996).
25. S. Mukhopadhyay, N.K.Das, A.Pradhan, N.Ghosh, P.K.Panigrahi, "Wavelet and multi-fractal based analysis on DIC images in epithelium region to detect and diagnose the cancer progress among different grades of tissues", Proc. of SPIE, SPIE Photonics Europe, pp. 91290Z-91290Z-7 (2014).
26. D. Donoho, I. Johnstone, G. Kerkyacharian, D. Pichard, "Wavelet shrinkage: Asymptopia?," J. Roy. Statist.Soc., vol. 57, pp. 301–369 (1995).
27. S.Mukhopadhyay, S.Mandal, N.K.Das, S.Dey, A.Mitra, N.Ghosh, P.K. Panigrahi, "Diagnosing Heterogeneous Dynamics for CT Scan Images of Human Brain in Wavelet and MFDFA domain", Proc. Of Springer, IEM Optronix-2014, India. (Accepted)